\pdfoutput=1

\documentclass[11pt]{article}

\usepackage{acl}

\usepackage{times}
\usepackage{latexsym}
\usepackage{amssymb}
\usepackage{amsmath}
\usepackage{booktabs}
\usepackage{graphicx}
\usepackage{subcaption}
\usepackage{xcolor}

\usepackage[T1]{fontenc}

\usepackage[utf8]{inputenc}

\usepackage{microtype}

\title{Combining Deep Neural Reranking and Unsupervised Extraction for Multi-Query Focused Summarization}

\author{Philipp Seeberger \and Korbinian Riedhammer \\
  Technische Hochschule Nürnberg Georg Simon Ohm \\
  \texttt{\{philipp.seeberger,korbinian.riedhammer\}@th-nuernberg.de}}

\begin{document}

\maketitle
\begin{abstract}

The CrisisFACTS Track aims to tackle challenges such as multi-stream fact-finding in the domain of event tracking; participants' systems extract important facts from several disaster-related events while incorporating the temporal order.
We propose a combination of retrieval, reranking, and the well-known Integer Linear Programming (ILP) and Maximal Marginal Relevance (MMR) frameworks.
In the former two modules, we explore various methods including an entity-based baseline, pre-trained and fine-tuned Question Answering systems, and ColBERT.
We then use the latter module as an extractive summarization component by taking diversity and novelty criteria into account.
The automatic scoring runs show strong results across the evaluation setups but also reveal shortcomings and challenges.

\end{abstract}

\begin{figure*}[t]
    \includegraphics[width=\linewidth]{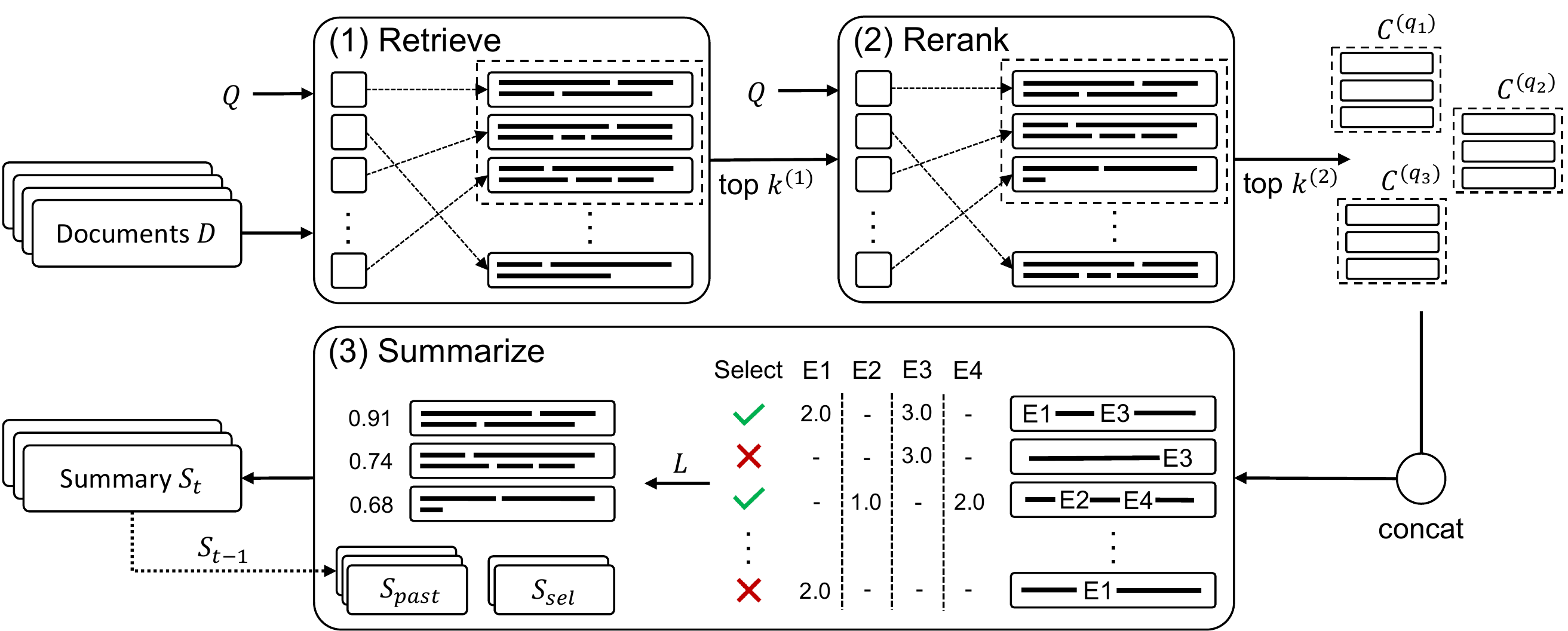}
    \caption{Overview of our proposed framework. All queries and documents for each event and time period are separately processed by the following three major components: (1) Retrieve, (2) Rerank, and (3) Summarize. The symbols E1 to E4 represent the concepts w.r.t. the ILP formulation. For final scoring, the selected $\mathcal{S}_{sel}$ and past summary $\mathcal{S}_{past}$ documents are used in terms of redundancy penalization.} 
    \label{fig:methods}
\end{figure*}

\section{Introduction}

Natural and human-made disasters can result in significant loss of life, property, and environment if situational awareness is insufficient due to a lack of critical information in an ongoing emergency event.
Today's information ecosystem offers new opportunities and directions for emergency response by integrating various online information sources \citep{trecis_2021,kruspe_2021}.
Additional information sources such as social media and microblogging platforms can immediately provide details about current developments \citep{sakaki_2010,reuter_2018}.
This leads to a multi-stream setting in which traditional sources are complemented with a variety of recently emerged online sources.
Previous research efforts acknowledged this setting as a promising venue, as shown by the evolving tasks over several decades \citep{allan_1998,aslam_2015,sequiera_2018,trecis_2021}.

However, the high-velocity nature of content generation and the inherent properties of different information sources \citep{kaufhold_2021} and events \citep{seeberger_2022} face present models with new challenges. Those provide relevant event-related results to the user but are still ill-suited to multi-stream fact-finding and summarization needs.
The novel CrisisFACTS Track aims to tackle these issues and challenges the community to develop systems more suited for factoid extraction over time.
Overall the task asks participants' systems to extract a query-focused list of facts from crisis-related datasets, including \textit{Twitter}, \textit{Reddit}, \textit{Web News}, and \textit{Facebook} as data sources.
Each of these extracted lists of facts is based on an event-day pair and shipped with importance scores which serve as a basis for downstream summarization.
In fact, this can be considered as an extension of previous tasks in the area of Information Retrieval (IR) and summarization.

The recent incorporation of pre-trained language models such as BERT \citep{devlin_2019} has significantly improved ad-hoc ranking \citep{lin_2022} and summarization \citep{ma_2022} results.
In particular, BERT-based cross-encoders achieve notable improvements over classical retrieval and dual-encoder approaches but have infeasible computational costs \citep{khattab_2020}.
To mitigate this issue, deep neural ranking models are typically deployed as second stage rerankers, whereby the first stage often represents an efficient retriever to create a subset of candidate documents \citep{macavaney_2022}.
Similarly, modern summarization methods rely on cross-encoders to fetch relevant documents, paragraphs, or sentences for both extractive summarization \citep{xu_2020,ahuja_2022} or as preliminary selection for abstractive summarization \citep{xu_2021}.
Finally, the resulting pool of ranked documents can be further refined through approaches such as MMR \citep{carbonell_1998}, ILP \citep{mcdonald_2007}, and TextRank \citep{mihalcea_2004}.

In this work, we explore various information retrieval and reranking pipelines ranging from pre-trained to fine-tuned state-of-the-art models. 
Complementary, we propose to subsequently process the list of facts in an extractive summarization setup by leveraging a combination of the well-known ILP and MMR frameworks.
In this way, we aim to overcome issues related to diversity and redundancy.

\section{Approach}

As illustrated in \figurename~\ref{fig:methods}, our proposed framework first retrieves and reranks a set of documents $\mathcal{D}=\{d_1,\ldots,d_{N}\}$ based on an information request $\mathcal{Q}=\{q_1,\ldots,q_{M}\}$ where each query $q_{i}$ typically consists of a short text or list of indicative terms.
The number of documents\footnote{Throughout this work, we use the term document interchangeably with the CrisisFACTS stream items. These stream items are rather sentences or short posts than long documents.} and queries are given as $N$ and $M$, respectively.
Let $\mathcal{C}=\{\mathcal{C}^{(q_1)},\ldots,\mathcal{C}^{(q_M)}\}$ denote the resulting set of query-related clusters with $\mathcal{C}^{(q_i)}=\{d_1^{(q_i)},\ldots,d_k^{(q_i)}\}$ consisting of the top $k$ candidates ranked by relevance.
Then, a summarization component further selects $L$ candidates from the cluster pool  $\bigcup_{i=1}^{M} \mathcal{C}^{(q_i)}$ to create a summary $\mathcal{S}$.
Following the track design, each summary $\mathcal{S}_{t}$ is created w.r.t. a time period $p_{t} \in \{p_{1},\ldots,p_{T}\}$ with $T$ as the number of time periods.
Within the scope of CrisisFACTS, the time period $p_{i}$ corresponds to one day.
In the following, we detail each individual component used for our submissions.

\subsection{Stage 1: Retrieve}

In the first stage, we employ lexical retrieval approaches to mitigate the infeasible computational costs of deep neural models such as BERT-based cross-encoders.
Hence, we first retrieve the top $k^{(1)}$ candidates for each query $q \in \mathcal{Q}$ from a set of documents $\mathcal{D}$ using a list of indicative terms.
For each query $q$, the reduced set of $k^{(1)}$ candidates is then subsequently processed by the given reranking stage.
For retrieval, we adopt the well-known BM25 model \citep{robertson_2009} but one can easily replace it with more sophisticated methods.
We empirically found at preliminary experiments that the number of candidates is relatively low, limiting the subsequent reranking components.
To address this problem, one can implement query expansion \cite{amati_2002}, document expansion \cite{nogueira_2019}, or adaptive reranking \citep{macavaney_2022} methods.
In this work, we integrate query expansion in order to grow the candidates pool for overcoming the recall limitation.

\subsection{Stage 2: Rerank}

Classical retrieval approaches may not be sufficient to capture the semantics expressed in the query and its relation to the text contents.
Therefore, each query-related cluster $\mathcal{C}^{(q)}$ of the previous retrieval stage is reranked and the top $k^{(2)}$ candidates are selected, resulting in a new set $\mathcal{C}'^{(q)}$.
Here, we exploit supervision signals from other existing datasets by using pre-trained deep neural models.
Furthermore, we include an entity and keyword-based baseline in order to assess the performance gain due to the models pre-trained on large text corpora.
In the following, we detail the considered reranking models:

\paragraph{BoE}
As baseline we adopt the Bag-of-Entities representation for document ranking introduced by \citet{xiong_2016} which relies on the semantic information achieved by entity linking systems.
We implement a simplified version by constructing Bag-of-Entities vectors for each document based on entity-types and extend them with Bag-of-Keywords vectors.
The final model scores a document by summing over the frequency of expected query entity-types and keywords present.

\paragraph{QA}
Similar to \citet{xu_2020}, we employ Question Answering (QA) systems to leverage distant supervision signals related to best answer selection.
While QA approaches support both sentence and span selection, we rely only on sentence level selection which suits more to the queries and stream items provided by the organizers.
That is, we concatenate the query $q$ and a candidate document $d$ into a sequence $\textnormal{[CLS]}~q~\textnormal{[SEP]}~d$ and predict the relevance score with a BERT-based cross-encoder.
In this work, we consider a pre-trained and fine-tuned QA version whereby the fine-tuned system is adapted to the crisis domain.

\paragraph{ColBERT} This model follows a contextualized late interaction approach which makes use of both a first-stage approximate nearest neighbor search and a reranking stage to calculate ranking scores \citep{khattab_2020}.
In particular, ColBERT supports the reranking mechanism to produce more precise scores based on a candidate pool but can also be used for end-to-end retrieval.
We follow the end-to-end retrieval approach.
In this way, we consider an approach without the limitations related to classical retrieval models such as BM25.
Note that this alternative skips the first retrieval stage by directly retrieving the set of documents $\mathcal{D}$ to obtain the top $k^{(2)}$ candidates for each query $q$.

\begin{table*}[t]
    \centering
    \begin{tabular}{lccccc}
        \toprule
        \multicolumn{1}{c}{} & \multicolumn{3}{c}{\textbf{Summarization}} & \multicolumn{2}{c}{\textbf{Matching}} \\
        \cmidrule(rl){2-4} \cmidrule(rl){5-6}
          & \textbf{ICS} & \textbf{NIST} & \textbf{Wiki} & \textbf{Comprehensiveness} & \textbf{Redundancy} \\
        \hline
        ColBERT & .050/\textbf{.450} & .139/.546 & .031/.542 & .189 & .201  \\
        BM25 $\rightarrow$ BoE & .047/.436 & .142/.560 & .030/.533 & .185 & \textbf{.176} \\
        BM25 $\rightarrow$ QA\textsubscript{ASNQ} & \textbf{.051}/.448 & \textbf{.147}/.563 & \textbf{.036}/\textbf{.565} & \textbf{.213} & .226 \\
        BM25 $\rightarrow$ QA\textsubscript{Crisis} & .046/.443 & \textbf{.147}/\textbf{.564} & .034/.545 & .210 & .226 \\
        \midrule
        TREC best & .058/.459 & .147/.564 & .036/.565 & .217 & .125 \\
        \bottomrule
    \end{tabular}
    \caption{Overall results of our automatic submission runs. We report the Rouge-2/BERTScore for summarization and comprehensiveness and redundancy ratio for matching as defined in \appendixname~\ref{appendix:metrics}. The top results across our proposed systems are in bold.}
    \label{table:results:overall}
\end{table*}

\begin{table}[t]
    \begin{tabular}{lccc}
        \toprule
       \textbf{Event} & \textbf{ICS} & \textbf{NIST} & \textbf{Wiki} \\
        \hline
        001 & .116$^{\star}$/.522 & \textbf{.273}/.560$^{\star}$ & .013$^{\star}$/.540 \\
        002 & \textbf{.066}/.561 & .050$^{\star}$/.563 & .043/\textbf{.579} \\
        003 & .053/.516 & .238/.611$^{\star}$ & .021/.593$^{\star}$ \\
        004 & .061/.480 & .171$^{\star}$/ \textbf{.585} & \textbf{.060}/\textbf{.582} \\
        005 & - & .136/\textcolor{gray}{.544}$^{\star}$ & .032/\textcolor{gray}{.526}$^{\star}$ \\
        006 & .057/.506 & .048/.533 & .019$^{\star}$/.580 \\
        007 & .040/\textcolor{gray}{.494}$^{\star}$ & .104$^{\star}$/.524$^{\star}$ & \textbf{.057}/\textbf{.554} \\
        008 & .012/.501 & \textbf{.154}/.583 & \textbf{.044}/\textbf{.562} \\
        \bottomrule
    \end{tabular}
    \caption{Rouge-2/BERTScore results for each event w.r.t. the QA\textsubscript{ASNQ} system. The TREC best results across all submissions are in bold. Results below the TREC median are in grey, while results below our baseline BoE are marked with $^{\star}$.}
    \label{table:results:summarization}
\end{table}

\subsection{Stage 3: Summarize}

\paragraph{Selection}
Finding a diverse set of facts with less redundancy is crucial for summarization tasks.
However, without any post-processing, reranked candidates still suffer in terms of diversity and redundancy.
To tackle this problem, we use an additional selection step formalized as ILP.
We follow the concept-based model \citep{gillick_2009,riedhammer_2010} where concepts can be facts, events, or information units.
In this problem setup, the objective function is maximized over the weighted sum of the concepts present in the selection, subject to a length constraint.
Finally, we obtain an extractive summary $\mathcal{S}_{t} = \{d_{1},\ldots,d_{l}\}$ where $|\mathcal{S}_{t}|$ is limited by $l \leq L$ with $L$ as the maximum number of documents.

\paragraph{Scoring}
The well-known MMR algorithm greedily selects documents by trading off query-based relevancy and redundancy to the previously selected documents, until a summary length constraint is met.
However, this constraint can be relaxed to rerank a summary $\mathcal{S}_{t}$ in order to increase the diversity in the top documents.
Formally, we define the final score of a document $i$ as
\begin{equation}
\lambda \cdot \textnormal{Rel}_{i} - (1 - \lambda) \cdot \max_{j \in \mathcal{S}_{sel} \cup \mathcal{S}_{past}} \textnormal{Red}_{ij}
\end{equation}
where Rel$_{i}$ is the relevance score of document $i$ and Red$_{ij}$ is the redundancy penalty for having both documents $i$ and $j$ in the summary $\mathcal{S}_{sel}$ as well as past summaries $\mathcal{S}_{past}=\bigcup_{i=1}^{t-1} \mathcal{S}_{i}$.
However, a single retrieved document might contain multiple scores due to multiple matched queries.
We argue that a document that covers multiple queries expresses more relevant information content for the summary.
Formally, we denote the relevance score as $\textnormal{Rel}_{i}=|\mathcal{Q}^{(i)}| \cdot score_{i}$ where $score_{i}$ is the mean score of document $i$ weighted by the number of matched queries $\mathcal{Q}^{(i)} \subseteq \mathcal{Q}$.

\section{Experiments}

In this section, we detail the experimental setup and discuss the results for our submitted runs. 
Throughout all experiments, we mainly consider the sources \textit{Twitter}, \textit{Reddit}, \textit{Web News} and ignore \textit{Facebook} due to the limited access to the post contents.

\subsection{Preprocessing}

We normalize all tweets in order to represent the text content similar to the other online sources.
Specifically, all retweet-indicating prefixes, user mentions, emoticons, emojis, and URLs are removed.
Furthermore, we remove any hashtag symbols and split the text into their corresponding words using \textit{WordSegment}.\footnote{\url{https://grantjenks.com/docs/wordsegment}}

\subsection{Crisis-QA}

Since the first CrisisFACTS Track does not provide any annotations w.r.t. the task, we decided to create a synthetic version that reflects the query-focused sentence selection.
We leverage the DocEE dataset \citep{tong_2022}, a recently published benchmark for document-level event extraction.
We extract a subset of 6818 documents which only covers crisis-related events and their corresponding event arguments.\footnote{We checked for an overlap between the DocEE and CrisisFACTS events. In fact, some of the events are part of the DocEE dataset and thus we removed the corresponding documents prior to our experiments.}
First, we manually create coarse-grained questions for each event argument.
Second, the dataset is augmented with a T5\textsubscript{BASE} question generation model\footnote{\url{https://huggingface.co/mrm8488/t5-base-finetuned-question-generation-ap}} for obtaining fine-grained questions.
Last, we synthesize question-sentence pairs based on the argument position and label this pairs as binary relevance classification task.
For model validation, we use the published dataset splits.

\subsection{Experimental Setup}

\paragraph{Retrieve}
For the first stage, we use the BM25 model with default settings of the PyTerrier library \cite{macdonald_2020} and extend it with the Bo1 query expansion \cite{amati_2002} component.
For each query, we concatenate the query text and indicative terms, retrieve the top $k^{(1)}=100$ candidates, and drop exact duplicates.
The majority of duplicates appear in the tweet documents which is mostly related to retweets.

\paragraph{Rerank}
The BoE model is based on a manually curated set of entity-types that mostly fits the expected information needs w.r.t. each query.
For example, queries about missing peoples typically cover numbers and locations, respectively.
The indicative terms provided by the organizers are used for the keywords.
The QA\textsubscript{ASNQ} system is based on RoBERTa\textsubscript{BASE} pre-trained on the ASNQ dataset \citep{garg_2020} without any further adjustments. 
Similarly, we employ the ColBERT\textsubscript{v2} version \citep{santhanam_2022} which is trained on the MS MARCO Passage Ranking task.
In terms of QA\textsubscript{Crisis}, we follow the adaptation step of \citet{garg_2020} by fine-tuning the QA\textsubscript{ASNQ} model on the domain-specific Crisis-QA dataset.
This results in an adapted version of the QA system.
Although the synthesized dataset relies on a broad range of labeled event arguments, we still observe a significant proportion of false negatives within the question-sentence pairs.
Hence, we use the model QA\textsubscript{Crisis-0} in a first step to denoise the dataset with an upper threshold of $0.1$ and then train a new model QA\textsubscript{Crisis-1} in a second step, which is in line with previous work such as RocketQA \citep{rocketqa_2021}.
We use the Transformers library \cite{wolf_2020} for the QA models, the official implementation of ColBERT, and select the top $k^{(2)}=25$ candidates for each query.

\paragraph{Summarize}
To enable a fair comparison among the different retrieval and reranking components, we re-use the selection and scoring procedure for each run.
Specifically, inspired by information extraction \cite{rodriguez_2020}, we extract entities\footnote{\url{https://stanfordnlp.github.io/stanza/ner.html}} as concepts, entity-frequency as weights, and set $L=150$ for the ILP formulation.
For MMR, we select $\lambda=0.8$ and calculate the redundancy $\textnormal{Red}_{ij}$ based on TF-IDF features and cosine similarity.

\subsection{Results}

In \tablename~\ref{table:results:overall}, we present the overall performance of our pipeline setups.
Since this is the first installment of the CrisisFACTS Track, we mainly limit the analysis across our submission runs.
However, we provide the reader a comparison of our models to the medians and top results for the summarization task (\tablename~\ref{table:results:summarization}).

\paragraph{Overall}
The QA models outperform the baseline BoE and ColBERT in almost all evaluation settings.
These results reflect the findings of previous text retrieval work which report higher performance for cross-encoder architectures.
Interestingly, the fine-tuned QA model decreases the performance in two summarization setups and in terms of comprehensiveness.
We assume that the adapted QA is biased towards the entities of the Crisis-QA dataset.
This might result into higher scores for only a subset of facts.
Furthermore, we are aware of concerns about potential data overlap due to the time intersection between the CrisisFACTS and Crisis-QA events.
However, the performance increase appears only for the NIST reference summaries and we therefore leave the analysis for future work.

\paragraph{Summarization}
In depth analysis in \tablename~\ref{table:results:summarization} found that the pre-trained QA model achieves top results for a variety of events and reference summaries.
When compared to the BoE baseline, the performance increase differs among the events, metrics, and reference summaries.
However, only three performance measures are below the TREC medians which suggests strong results for the overall pipeline.
Nevertheless, in contrast to automatic summarization evaluation, manual matching reveals high variance for different days within the same event.

\begin{figure}[t]
    \centering
    \includegraphics{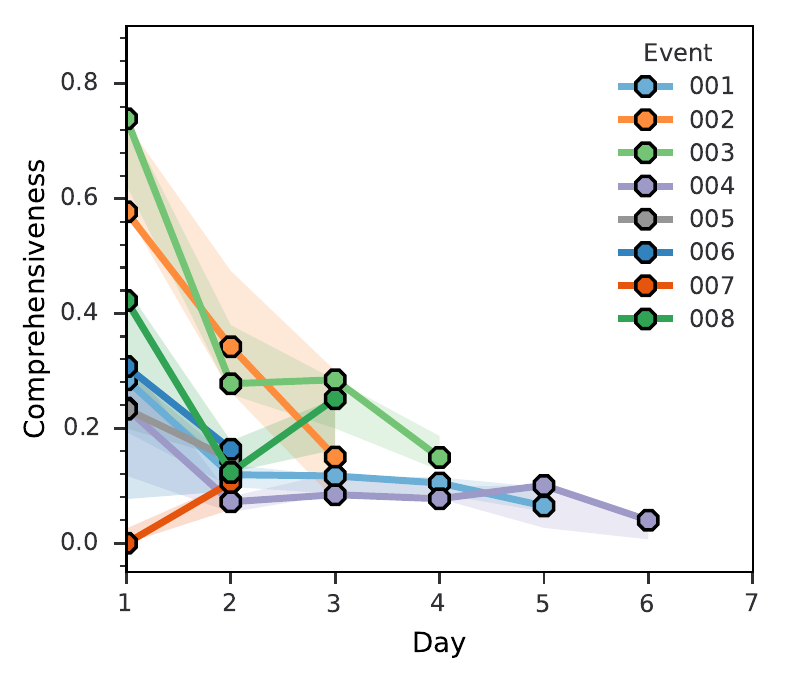}
    \caption{Comprehensiveness trend for all events. The QA\textsubscript{ASNQ} system is displayed in bold, while the min-max region of all models is highlighted.}
    \label{fig:results:trend}
\end{figure}

\paragraph{Matching}
If we plot the comprehensiveness evolution along the number of days (\figurename~\ref{fig:results:trend}), we see that the performance decreases by large extent across a variety of events.
Since this trend holds for all models, we hypothesize that this is due to at least two factors.
First, the retrieval and reranking stages of the pipeline setup does not consider diversity for each query and might cut off rare facts in favor of facts with higher relevance, spread along the timeline of the event.
Second, the diversification in the selection stage w.r.t. past summaries still displays a challenging task.
For example, specific sentences only differ by a single number (e.g. burned acres) and might unintentionally penalize new facts by unsophisticated similarity measures.

\section{Conclusion}

In this work, we have investigated the combination of deep neural reranking and global unsupervised extraction for a multi-query focused summarization task.
Our experiments demonstrated the strength of cross-encoders with QA based on distant supervision.
However, we identified shortcomings and challenges in the face of temporal aspects which underlines the downstream summarization as a critical component.
We believe there is much room for improvement, especially by integrating more sophisticated extractive approaches, abstractive summarization techniques, or even joint optimization. 

\section*{Acknowledgments}

The authors acknowledge the financial support by the Federal Ministry of Education and Research of Germany in the project ISAKI (project number 13N15572).

\bibliography{anthology,custom}
\bibliographystyle{acl_natbib}

\appendix

\section{Matching Metrics}
\label{appendix:metrics}

The submitted stream items for a system and event-day pair are ordered by the importance scores and formed to a summary $\mathcal{S}$ by a rank cut-off $k$. 
Based on a manual matching of the stream items against a fact list $F$, the comprehensiveness is calculated as

\begin{equation}
\frac{1}{\sum_{f \in F} R(f)} \sum_{\{f \in F:M(f,\mathcal{S}) \neq \emptyset \}} R(f)
\end{equation}

where $f \in F$ is a unique fact, $M(f,\mathcal{S})$ is the set of stream items matching fact $f$, and $R(f)$ is the gain assigned to the fact $f$. 
Similarly, the redundancy ratio is measured for a system and event-day pair as

\begin{equation}
\frac{\sum_{\{f \in F:M(f,\mathcal{S}) \neq \emptyset\}} R(f)}{\sum_{\{f \in F\}} R(f) \cdot |M(f,\mathcal{S})|} 
\end{equation}

All runs are macro-averaged across days within an event, and then across the eight events.

\end{document}